\pdfoutput=1
\documentclass[11pt]{article}

\usepackage[preprint]{acl}

\usepackage{times}
\usepackage{CJKutf8}
\usepackage{latexsym}

\usepackage[T1]{fontenc}
\usepackage[utf8]{inputenc}

\usepackage{microtype}

\usepackage{inconsolata}

\usepackage{graphicx}

\usepackage{booktabs}
\usepackage{colortbl}
\usepackage{tabularx}
\usepackage{multirow}
\usepackage{array}
\usepackage{subcaption}
\usepackage{amssymb}
\usepackage{longtable}
\usepackage{color,xcolor}
\definecolor{darkgreen}{rgb}{0.0, 0.6, 0.0}

\newcommand{\method}{\textsc{RecKon}}
\definecolor{mygray}{gray}{0.9}
\title{\method: Large-scale \underline{Re}ference-based Effi\underline{c}ient \underline{K}nowledge Evaluati\underline{on} for Large Language Model}

\author{
Lin Zhang\textsuperscript{\rm $\spadesuit$}\thanks{Equal Contribution},
Zhouhong Gu\textsuperscript{\rm $\spadesuit$*},
Xiaoran Shi\textsuperscript{\rm $\spadesuit$},
Hongwei Feng\textsuperscript{\rm $\spadesuit$}\thanks{Corresponding authors.}\ ,
Yanghua Xiao\textsuperscript{\rm $\spadesuit$$\dagger$}\\
\textsuperscript{\rm $\spadesuit$}Shanghai Key Laboratory of Data Science, School of Computer Science, Fudan University\\
\{linzhang22, zhgu22, xrshi21\}@m.fudan.edu.cn\\
\{hwfeng, shawyh\}@fudan.edu.cn
}

\begin{document}
% \begin{CJK}{UTF8}{gbsn}
\maketitle
\begin{abstract}
As large language models (LLMs) advance, efficient knowledge evaluation becomes crucial to verifying their capabilities. Traditional methods, relying on benchmarks, face limitations such as high resource costs and information loss. We propose the Large-scale \underline{Re}ference-based Effi\underline{c}ient \underline{K}nowledge Evaluati\underline{on} for Large Language Model (\method), which directly uses reference data to evaluate models. \method~organizes unstructured data into manageable units and generates targeted questions for each cluster, improving evaluation accuracy and efficiency. Experimental results show that \method~reduces resource consumption by 56.5\% compared to traditional methods while achieving over 97\% accuracy across various domains, including world knowledge, code, legal, and biomedical datasets.
Code is available at \url{https://github.com/MikeGu721/reckon}.
\end{abstract}

\section{Introduction}
% As large language models (LLMs) continue to advance rapidly, knowledge evaluation has emerged as a crucial component for verifying their capabilities and driving systematic improvements. 
% The fundamental objective of evaluation is to assess and ensure the alignment between model performance and reference data. 
% Comprehensive and accurate knowledge evaluation is particularly vital in ensuring reliable model performance across diverse domains, with special emphasis on high-stakes fields such as healthcare, education, and law, where robust evaluation methodologies are imperative for mitigating risks associated with errors or outdated information. 
% Given the accelerating proliferation of LLM applications, the scope and complexity of knowledge evaluation requirements have expanded considerably. 
% Contemporary evaluation frameworks must address increasingly demanding criteria, including broader domain coverage, more sophisticated analytical tasks, and the processing of substantially larger datasets, collectively presenting unprecedented challenges to traditional evaluation methodologies.

As large language models (LLMs) continue to advance rapidly, knowledge evaluation has become an essential component for verifying their capabilities and driving continuous improvements.
The fundamental objective of evaluation is to assess and ensure the LLM's alignment with reference data.
Thorough and accurate knowledge evaluation plays a critical role in ensuring reliable model performance across different fields, especially in crucial areas such as healthcare, education, and law, where strong evaluation methods are necessary to reduce risks from errors or outdated information. 
With the rapid growth of LLM applications, the requirements for knowledge evaluation have become increasingly extensive.
Modern evaluation approaches must meet growing demands, including wider coverage of different fields, more complex analytical tasks, and the handling of much larger datasets. 
These requirements pose significant challenges to traditional evaluation methods.

\begin{figure}[t]
    \centering
    \includegraphics[width=0.99\linewidth]{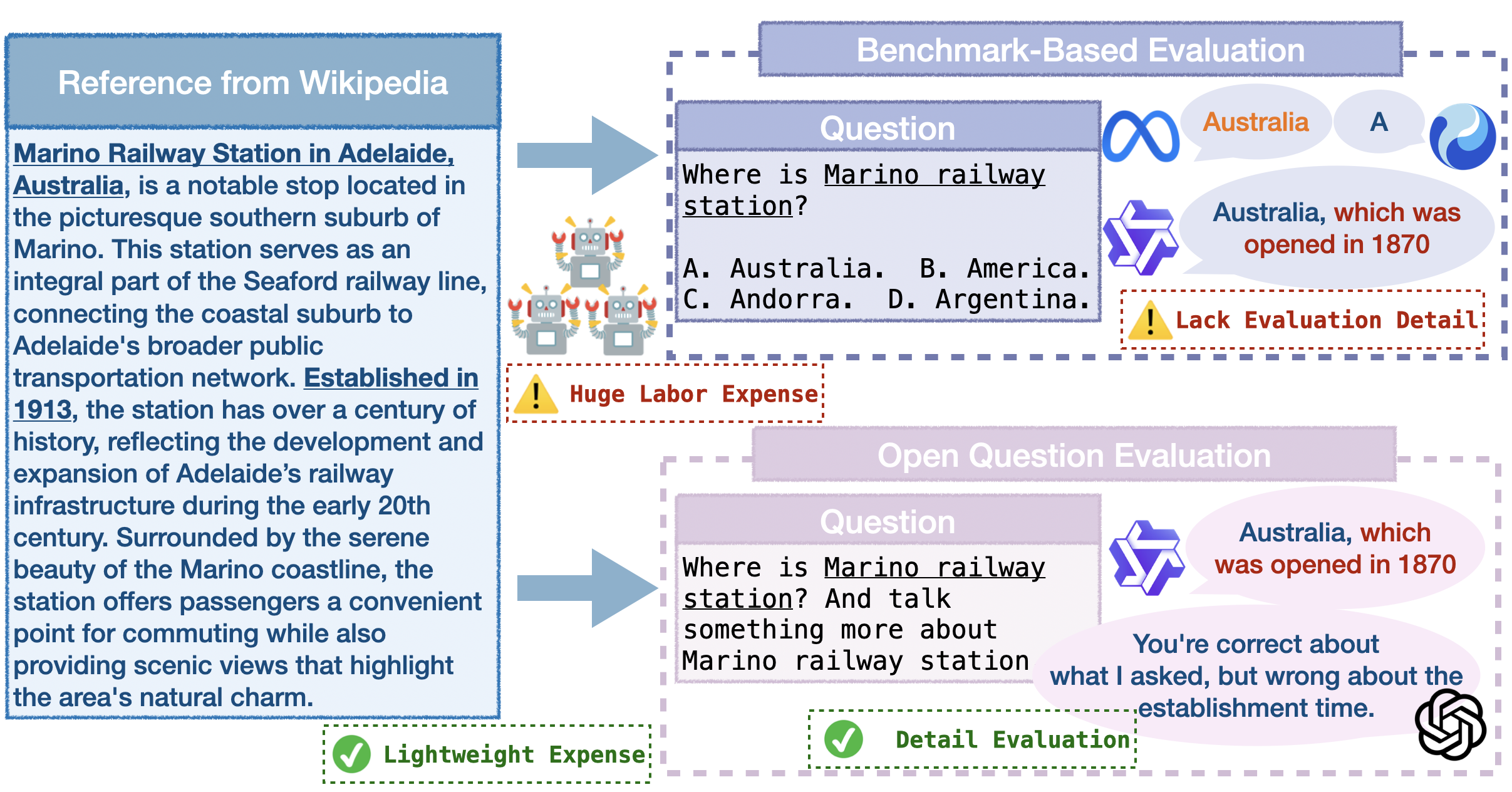}
    \caption{The comparison of different evaluation method.}
    \label{fig:intro}
\end{figure}

Traditional knowledge evaluation methodologies primarily rely on benchmarks, encompassing various assessment formats such as multiple-choice questions~\cite{gu2024xiezhi}, fill-in-the-blanks exercises~\cite{hendrycks2020measuring}, and marking tasks~\cite{dimitrov2024semeval}. 
These benchmarks fundamentally serve as condensed or refined representations of reference data, carefully designed to facilitate manageable and systematic evaluation processes. 
Despite widespread adoption across the field, these benchmark-based approaches exhibit several inherent limitations. 
First, the creation and verification of comprehensive benchmarks demand substantial resources, necessitating significant manual and computational expenses. 
Second, the process of distilling complex, multifaceted information into standardized questions inevitably results in information loss, potentially introducing systematic biases and compromising the validity of knowledge assessment. 
Finally, individual benchmarks face considerable challenges in fully capturing the intricate complexity and expansive diversity of knowledge domains, thereby constraining their adaptability and broader applicability.

As illustrated in Fig.\ref{fig:intro}, directly leveraging rich and dynamic reference data is an alternative approach, which presents distinct advantages over traditional benchmark-based methods. 
This direct reference approach not only preserves the nuanced complexity of the original information but also enables more comprehensive and authentic evaluation scenarios. 
Furthermore, it facilitates dynamic updating of evaluation criteria as new knowledge emerges, ensuring continued relevance and accuracy. 
The approach also demonstrates superior scalability across different knowledge domains and evaluation contexts, offering enhanced flexibility in evaluation. 
However, this direct reference-based evaluation introduces its own set of significant challenges: 
First, reference datasets typically exist in large-scale, unstructured formats, presenting substantial difficulties in effective organization and utilization during evaluation processes. 
Second, the absence of predefined questions or standardized answer formats complicates the establishment of consistent and equitable alignment between model outputs and reference data.

To address these challenges, we propose a \underline{Re}ference-based Effi\underline{c}ient \underline{K}nowledge Evaluati\underline{on} framework, called \method, which directly and efficiently evaluates LLMs using reference data. 
In response to the difficulty of organizing and utilizing large and unstructured reference datasets, \method~decomposes the reference data into discrete, manageable knowledge units. 
These units are then grouped into thematic clusters, forming an organized knowledge structure that streamlines the evaluation process. 
To solve the problem of ensuring consistent and fair alignment without predefined questions or fixed answers, \method~generates targeted questions for each knowledge cluster.
The evaluation focuses on assessing whether the model’s responses adequately cover all relevant knowledge units, ensuring comprehensiveness and accuracy.

Experimental results highlight the superiority of \method~ as a comprehensive, efficient, and adaptable framework for knowledge evaluation.
Across various common-sense and anti-common-sense tasks, \method~effectively adapts references to reduce the inherent biases of LLMs.
Notably, \method~ reduced resource consumption by 56.5\% compared to ``full reference input evaluation'' without compromising evaluation accuracy.
Additionally, \method~ demonstrated exceptional adaptability across diverse domains, including world knowledge, code, legal, and biomedical datasets, achieving an accuracy exceeding 97\%. 

\section{Related work}
Open-ended question answering benchmarks for models include MMLU~\cite{hendrycks2020measuring}, C-eval~\cite{huang2024c}, Xiezhi~\cite{gu2024xiezhi}, etc. 
These benchmarks contain a series of questions that are described in natural language and require the model to give an open-ended answer.
However, these benchmarks rely on a large number of manual annotations and cannot be updated with the latest knowledge. 
Currently, some methods propose using LLMs to automatically build updatable benchmarks, such as LM-as-an-Examiner~\cite{bai2024benchmarking} and TreeEval~\cite{li2024treeeval}. In these methods, when no benchmark is available, LLMs generate questions based on their own knowledge.
However, the knowledge in LLM cannot be inherently complete, and the language model is biased, which will lead to incomplete questions and deviation of questions to a certain extent.

Traditional open evaluation metrics are based on n-grams to measure semantic similarity between texts, including BLEU~\cite{papineni2002bleu}, ROUGE~\cite{lin2004rouge}, METEOR~\cite{banerjee2005meteor}, etc. 
These methods enable the automated evaluation of natural language texts. 
However, these methods lack expressive power and cannot semantically distinguish key information, such as negative words.
To address these problems, BERTScore~\cite{zhang2019bertscore}, GPTScore~\cite{fu2023gptscore}, and other methods use language models to derive similarities between candidate answers and reference answers. 
In the latest evaluation work on LLM, LLM-as-Judge (a paradigm that uses LLM to evaluate open results)~\cite{zheng2024judging, chan2023chateval} is proposed. 
This method is recognized for its strong interpretability and scalability, making it widely adopted.
These methods still rely on either artificially constructed reference answers or the LLM's own knowledge, leaving the issue of knowledge limitations unresolved.

\method~distinguishes itself from traditional benchmarks by using a dynamic, reference-based approach rather than static question-answer pairs.  
This design makes it more adaptable, scalable, and effective at evaluating nuanced knowledge across diverse domains.  
Unlike traditional benchmarks, which are limited in coverage, flexibility, and efficiency, \method~leverages rich reference data to align with evolving knowledge.  
Moreover, \method~distinguishes itself from other LLM-related evaluation methods, such as LLM-as-Judge, by grounding evaluations in external references rather than relying on the model's subjective judgment. This ensures a more objective, transparent, and consistent evaluation process, effectively mitigating biases and variability inherent in LLM-as-Judge approaches.

\section{\method~Framework}\label{sec:method}
\begin{figure*}
    \centering
    \includegraphics[width=1\linewidth]{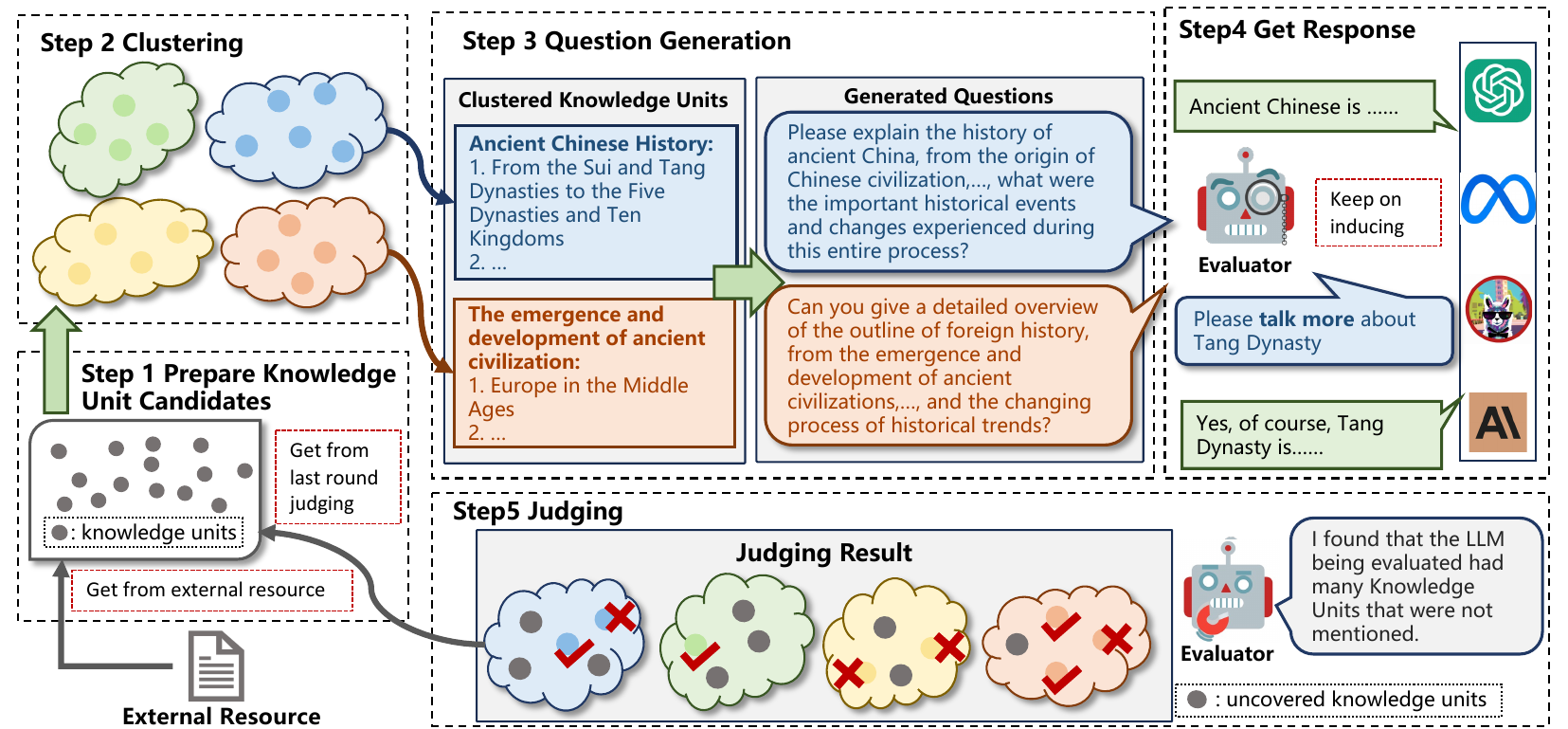}
    \caption{\method~consists of five main components that form a cycle of assessment.}
    \label{fig:method}
\end{figure*}

As shown in Fig.\ref{fig:method}, \method~consists of five main components, and the description of these components is as follows:

\textbf{Step 1 Preparing Knowledge Unit Candidates:}
The \method~ framework begins by preparing knowledge units that will be utilized throughout the evaluation process. These units are sourced primarily from two places: External references, such as curated datasets, textbooks, or knowledge bases relevant to the target domain. And Previously evaluated knowledge units, units from earlier rounds of assessment that were flagged as uncovered or incomplete by the evaluated model.

\textbf{Step 2 Clustering:}
After compiling the knowledge units, clustering step involves organizing them into meaningful clusters to streamline the evaluation process. 
Clustering effectively organizes knowledge units into coherent groups to facilitate the evaluation process. This organization is achieved by analyzing the semantic relationships between different knowledge units. Each knowledge unit, which may consist of facts, concepts, and information relevant to the related field, is transformed into an embedding representation. These units are then classified based on the proximity of their embeddings in semantic space, effectively creating subsets of knowledge that are thematically related.

\textbf{Step 3 Question Generation:}
In this step, \method~ employs a LLM to generate targeted questions for each cluster of knowledge units. 
This step utilizes the capabilities of the language model to pose questions.
The questions are designed to align with the central theme of the cluster, ensuring they are contextually appropriate and address the core knowledge units.
These questions are tailored to test not only surface-level understanding but also deeper insights, relationships, and nuances within the knowledge units.
During the evaluation, if the language model's responses reveal gaps, ambiguities, or misconceptions, \method~ dynamically adjusts the questions to probe these areas further. This iterative refinement ensures a comprehensive assessment of the evaluated model's knowledge and reasoning capabilities.

\textbf{Step 4 Get Response:}
Once the questions are generated, they are presented to get responses from the evaluated targets.
The responses obtained during this phase serve as the primary input for analysis.
\method~ evaluates the accuracy, completeness, and relevance of the responses in relation to the knowledge units.
If a response fails to address a question, \method~ adequately will continue to prompt the model to generate questions to cover the relevant knowledge units. 
This ensures that every relevant knowledge unit within the cluster is thoroughly examined.

\textbf{Step 5 Judging:}
This step involves instructing the evaluation LLM to assess the extent to which the evaluated model's responses cover the specified knowledge units. 
In this phase, \method~ labels the covered knowledge units as either correct or incorrect. 
For knowledge units that remain uncovered by the evaluated model's response, \method~identifies and retains them for reorganization and re-evaluation. 
The process starts again from the initial step in subsequent assessment rounds.

Further technical details, including the prompts, models, inputs, and outputs used at each step, can be found in Appendix \ref{A.1}.

% The final step involves a detailed evaluation of responses using a combination of automated and manual review methods to assess the correctness, relevance, and depth of the response provided by the language model. 
% This step is iterative, with each round of questioning and answer review refining the model's understanding of both known and unknown content.
% Through this continuous cycle, \method~identifies and retains unresolved knowledge units, prioritizing them in subsequent evaluations to progressively improve the accuracy and comprehension capabilities of the model.

\section{Experiment Setup}

\subsection{Dataset and models}

Our experiments encompassed diverse textual corpora spanning multiple domains:
\begin{itemize}
    \item \textbf{Wikitext} from \textbf{Wiki}~\cite{vrandevcic2014wikidata}, comprising meticulously curated articles encompassing comprehensive knowledge. The diversity and comprehensiveness of Wikitext make it an excellent benchmark for assessing the breadth and depth of a model’s general knowledge capabilities.
    \item \textbf{Code} from \textbf{CodeGPT}~\cite{codegpt}, containing programming-oriented assignments. It is specifically used to evaluate the models' ability to understand and generate code or solve programming challenges.
    \item \textbf{Legal.term} from \textbf{Legalbench}~\cite{guha2024legalbench}, incorporating juridical definitions emphasizing legislative interpretation. It is a resource for testing a model's performance in legal reasoning, terminology comprehension, and domain-specific text generation.
    \item \textbf{Med.rand} and \textbf{Med.sim} from \textbf{PubMedQA}~\cite{jin2019pubmedqa}. Med.rand includes randomized selections from the PubMedQA dataset to evaluate general biomedical knowledge. Med.sim is a similarity-based subset specifically designed to test the models’ ability to identify nuanced relationships between biomedical entities.
    \item \textbf{Uncommon} from \textbf{Feverous}~\cite{aly2021feverous}, encompassing descriptions that deliberately deviate from established factual paradigms. It assesses a model’s ability to handle the contradiction between an evaluation model and the external reference.
\end{itemize}

Details of these datasets is presented in Tab.\ref{tab:data}.

The models under evaluation include the widely used \textbf{GPT4-turbo} (GPT4), \textbf{GPT3.5-turbo} (GPT3.5), and \textbf{LLaMA2-chat-13b} (LLaMA2). 
Additionally, we incorporated the \textbf{PMC-LLaMA-13b}(PMC), fine-tuned on the LLaMA2 architecture, specifically tailored for the biomedical field, to assess its performance within the related domain.  

\begin{table}[t]
\centering
\small
% \resizebox{0.99\columnwidth}{!}{
\begin{tabular}{l|lrrc}
\toprule
\textbf{Dataset}             & \textbf{Source}     & \textbf{\#Words} & \textbf{\#Para.} \\
\midrule
Code       & CodeGPT\footnotemark    & 2,371k  & 5,511         \\
Legal.term& Legalbench\footnotemark & 233k   & 695           \\
Wikitext     & Wiki\footnotemark       & 11,834k & 4,396         \\
Uncommon     & Feverous\footnotemark   & 49k & 2,000         \\
Med.rand     & PubMedQA\footnotemark   & 877k   & 2,000         \\
Med.sim    & PubMedQA   & 351k   & 2,000         \\
\bottomrule
 \end{tabular}
 % }
\caption{
Detail statistic information of the datasets encompassed in the experiments.
}
\label{tab:data}
\end{table}

\subsection{Baseline}

In our evaluation, we compare \method~against several baseline methods: 
\textbf{BLEU}, a probabilistic measure based on n-gram matching by comparing the overlap of n-grams between two sentences.
\textbf{GPT Score}, an embedding-based indicator utilizing pre-trained language model embeddings to compare the semantic similarity between generated and reference texts.
\textbf{LLM-as-Judge w/o reference(Jw/oR)}, adopting the LLM-as-judge paradigm and relying on the evaluator (LLM) to judge response correctness.
\textbf{LLM-as-Judge w/ reference(Jw/R)}, assessing response correctness by aligning the answer provided by the model and the related textual context in the datasets one-by-one, which is superior to \method~with more detail.

\setcounter{footnote}{1}
\footnotetext{\url{https://github.com/zxx000728/CodeGPT/}}
\setcounter{footnote}{2}
\footnotetext{\url{https://hazyresearch.stanford.edu/legalbench/}}
\setcounter{footnote}{3}
\footnotetext{\url{https://huggingface.co/datasets/wikitext}}
\setcounter{footnote}{4}
\footnotetext{\url{https://github.com/Raldir/FEVEROUS}}
\setcounter{footnote}{5}
\footnotetext{\url{https://pubmedqa.github.io}}

% \begin{itemize}
%     \item \textbf{BLEU}, a probabilistic measure based on n-gram matching by comparing the overlap of n-grams between two sentences.
%     \item \textbf{GPT Score}, an embedding-based indicator utilizing pre-trained language model embeddings to compare the semantic similarity between generated and reference texts.
%     \item \textbf{LLM-as-Judge w/o reference(Jw/oR)}, adopting the LLM-as-judge paradigm and relying on the evaluator (LLM) to judge response correctness.
%     \item \textbf{LLM-as-Judge w/ reference(Jw/R)}, assessing response correctness by aligning the answer provided by the model and the related textual context in the datasets one-by-one, which is superior to \method~with more detail.
% \end{itemize}

% Furthermore, to establish ground truth for evaluating our metrics, we conducted \textbf{Human} evaluation as detailed in \ref{A.3}.
% Three annotators assessed model responses across all datasets with access to the original reference texts, scoring each response as either incorrect or correct.
% These human judgments serve as the gold standard for measuring the effectiveness of our automated evaluation methods.
To establish a ground truth for comparison, we conducted \textbf{Human} evaluations as detailed in Appendix \ref{A.3}. 
Three expert annotators assessed model responses across all datasets with access to the original reference texts. 
Each response was labeled as either correct or incorrect, based on its alignment with the reference text and overall quality.
The human evaluations served as the gold standard for our study, allowing us to measure the accuracy and reliability of \method~and other baseline methods. 
The average human scores and inter-annotator agreement are presented in \ref{A.3}.

\section{Performance}
Tab.\ref{tab:corr} displays the performance of all baselines measured by their correlation with human annotations.
Tab.\ref{tab:effi} displays the resource consumption of two strong baselines, Jw/R and \method. The conclusions from the results are as follows:

\paragraph{LLM-based methods exhibit significantly higher evaluation accuracy compared to both n-gram-based and embedding-based approaches.} 
The correlation with human baselines has notably improved, ranging from approximately 0.3 for BLEU to 0.6 for GPT Score and approaching nearly 1 for Jw/R, Jw/oR, and \method. 
This substantial enhancement underscores the superior efficacy of LLM-based techniques in assessing model responses in alignment with human expectations. 
The inferior performance of n-gram-based and embedding-based methods can be attributed to their limited understanding of the text's intrinsic meaning. 
In contrast, LLMs offer robust capabilities in comprehending textual context, thereby enhancing the effectiveness of Jw/oR methodologies.

\paragraph{External references enhance evaluation accuracy and stability by countering internal model biases.}
Firstly, references improve the accuracy of evaluation. 
Both the Jw/R and \method~baselines demonstrate a higher correlation, utilizing textual references to augment the LLM's precision in assessing answer correctness. 
External references effectively counteract the potential influence of erroneous internal model knowledge, as previously investigated in the literature \cite{xie2023adaptive}. 
Secondly, adopting external references leads to superior stability and robustness against LLM alone.
Notably, the Jw/R and \method~methods exhibit significantly lower variances of 1.9e-5 and 2.5e-4, respectively.
In contrast, the Jw/oR baseline demonstrates a higher variance of 0.04, indicating greater susceptibility to internal model biases.

\paragraph{\method~reduces the cost of evaluation and maintains competitive correlation with Jw/R.} 
Tab.\ref{tab:effi} demonstrates the efficiency advantage of \method. 
In contrast to Jw/R, which necessitates more frequent API calls and processes greater data volumes, thereby escalating operational expenses, \method~optimizes these aspects by concentrating on crucial tokens during the Judgment and Question generation phases.
This strategy reduces unnecessary overhead and lowers API call rates, enabling cost savings and effective evaluation.
A deeper analysis in \ref{sec:ana1} shows that \ method's cost is related to model capacity, with higher-capacity models incurring lower expenses compared to Jw/R.

\section{Analysis}
\begin{table}[t]
    \centering
    \small
% \renewcommand{\arraystretch}{4} % 设置表格行高
% \fontsize{16}{5.5}\selectfont

    \begin{tabular}{l|ccc}
    \toprule
        & \textbf{GPT4} & \textbf{GPT3.5} & \textbf{LLaMA2}\\
        \hline
         \multicolumn{4}{l}{\cellcolor{mygray}\textit{\textbf{N-Gram Based}}} \\
        \hline
        \addlinespace[2pt]
        \textbf{BLEU} & 28.1&38.4&9.4  \\
        
        \hline
         \multicolumn{4}{l}{\cellcolor{mygray}\textit{\textbf{Embedding Based}}} \\
        \hline
        \addlinespace[2pt]
        \textbf{GPTScore}&60.6&62.3&41.0\\

        \hline
         \multicolumn{4}{l}{\cellcolor{mygray}\textit{\textbf{LLM-as-Judge Based}}} \\
        \hline
        \addlinespace[2pt]
        \textbf{Jw/R}&\textbf{99.8}&98.9&\textbf{99.2}\\
        \textbf{Jw/oR}&88.9&98.4&86.1\\
        \textbf{\method}&97.6&\textbf{99.5}&96.3\\
    \bottomrule
    \end{tabular}
    % }
    
    \caption{Correlations between different methods and human baseline, where the bold font indicates the highest correlation.
    }
    \label{tab:corr}
\end{table}
\begin{table}[t]
\centering
\small
\begin{tabular}{p{0.49\textwidth}}
\toprule
\textbf{Question:} What was the Silent Holy Stones' director award in 2005? \\
\textbf{Answer:} Directed by Pema Tseden, won the Golden Rooster Award for Best Directorial Debut in 2005, marking a significant step for Tibetan cinema. \\
\textbf{Judgement:} \textcolor{red}{Incorrect} \\
\midrule
\textbf{Reference Text:} The Silent Holy Stones was nominated in the \textbf{Best Director category in the 2005 Beijing College Student Film Festival}. \\
\bottomrule
\end{tabular}
\caption{Performance of GPT-4 on Uncommon Dataset by \method.}
\label{tab:performance_ref_eval_uncommon}
\end{table}

\begin{table*}[ht]
\centering
% \small
% \renewcommand{\arraystretch}{2.7} % Set table row height
% \fontsize{14}{5.5}\selectfont
% \resizebox{0.99\textwidth}{!}{
\begin{tabular}{@{}l|r|r|r|r|r|r@{}} \toprule
& \textbf{Eval. Target}& \textbf{\#Query-API}  & \textbf{\#Token} & \textbf{\#Judge-API} & \textbf{\#Token} & \textbf{\#Money} \\
\midrule
\textbf{Jw/R} & GPT3.5 & 1,423 & 650,098 & 1,423 & 482,779 & 38.14 \\
\textbf{\method }& GPT3.5 & 256 & 26,880 & 1,622 & 616,760 & 2.44 \\
\textbf{\method} & GPT4 & 244 & 21,692 & 1,593 & 556,257 & 2.39 \\
\textbf{\method} & LLaMA & 269 & 69,271 & 1,850 & 660,350 & 3.27 \\
\bottomrule
\end{tabular}
% }
\caption{
Metrics of resource consumption measured during the Wiki evaluation phase.
\textbf{Eval. Target} signifies the language model subjected to assessment.
\textbf{\#Query-API} and \textbf{\#Judge-API} enumerate the aggregate API invocations throughout the question formulation and assessment phases, respectively.
\textbf{\#Token} encompasses the collective token consumption across all operations.
\textbf{\#Money} encapsulates the overall monetary expenditure incurred through API utilization.
}
\label{tab:effi}
% \vspace{-5mm}
\end{table*}

\begin{table}[t]
\centering
\small
\begin{tabular}{l|c|c|c|c}
\toprule
\textbf{Model}   & \textbf{GPT4} & \textbf{GPT3.5} & \textbf{LLaMA2} & \textbf{PMC} \\
\midrule
\textbf{Code}     & \textbf{95.4}       & 92.9                 & 83.1                & --                   \\
\textbf{Legal.term} & \textbf{94.6}       & 90.8                 & 81.2                & --                   \\
\textbf{Wiki}     & \textbf{82.6}       & 54.5                 & 42.9                & --                   \\
\textbf{Med.rand} & \textbf{95.3}       & 82.1                 & 73.2                & 78.5                \\
\textbf{Med.sim}  & \textbf{97.9}       & 82.3                 & 72.5                & 75.7                \\
\bottomrule
\end{tabular}
\caption{The performance of the model on each dataset, expressed as accuracy, where the bold font indicates the highest accuracy.}
\label{tab:main1}
\end{table}

\begin{figure*}[t]
    \centering
    \includegraphics[width=0.8\linewidth]{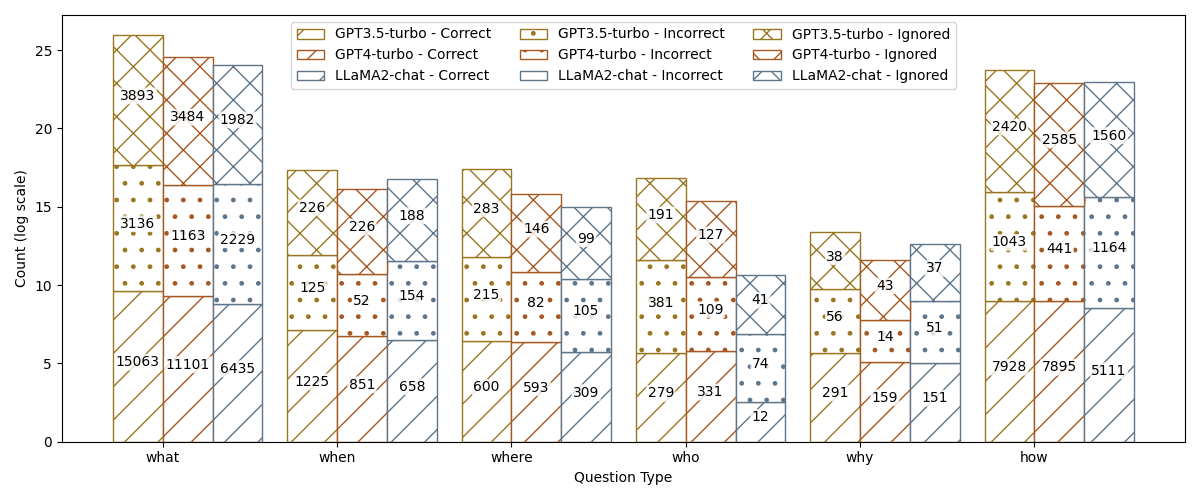}
    % \vspace{-3mm}
    \caption{The counts of 6 types of questions of \method~used for different models.}
    \label{fig:question_classification}
\end{figure*}

\begin{table}[]
\small
\centering
\begin{tabular}{@{}l|cccccc@{}}
\toprule
           & \textbf{Code}   & \textbf{Legal.term} & \textbf{Wiki}   & \textbf{Med.rand}  \\ \midrule
\textbf{Jw/oR} & 97.2  & 95.5     & 72.6 & 90.9    \\
\textbf{\method}  & 95.4 & 90.9     & 54.5  & 82.1    \\ 
\bottomrule
\end{tabular}
\caption{Evaluation result of \method~and raw LLM-as-a-Judge.}
\label{tab:wo}
\end{table}

\begin{table}[]
\centering
\small
\begin{tabular}{@{}lccc@{}}
\toprule
\multicolumn{4}{c}{\textbf{Model Accuracy}} \\ 
\cmidrule(l){2-4}
\multicolumn{1}{l|}{}           & \textbf{GPT4} & \textbf{GPT3.5} & \textbf{LLaMA2} \\ 
\cmidrule(lr){1-4}
\multicolumn{1}{l|}{\textbf{\method}}   & 31.7 & 32.8   & 24.7   \\ 
\multicolumn{1}{l|}{\textbf{Jw/oR}} & 84.4 & 87.3   & 67.5   \\ 
\midrule
\midrule
\multicolumn{4}{c}{\textbf{Correlation Between Human Annotation}} \\
\cmidrule(l){2-4}
\multicolumn{1}{l|}{}           & \textbf{GPT4} & \textbf{GPT3.5} & \textbf{LLaMA2} \\ 
\cmidrule(lr){1-4}
\multicolumn{1}{l|}{\textbf{\method}}   & 90.57 & 92.45 & 96.22\\
\bottomrule
\end{tabular}
\caption{Evaluation performance on UnCommon.}
\label{tab:uncomman}
\end{table}

\begin{table*}[t]
\centering
\renewcommand{\arraystretch}{2} % 设置表格行高

\fontsize{8}{5.5}\selectfont
% \small
\begin{tabularx}{\textwidth}
{@{}p{0.08\textwidth}|p{0.61\textwidth}|X@{}}
\midrule

\multicolumn{3}{l}{\cellcolor{mygray} \textbf{Comparison of \method~and LLM-as-Judge}} \\
% \midrule
\textbf{Jw/oR} &
\begin{tabular}{>{\raggedright\arraybackslash}m{0.6\textwidth}}
 \textbf{Question:} When was Marino railway station in Adelaide opened?  \\
 \textbf{Answer:} Marino Railway Station was opened on October 19, 1885. \\
 \textbf{Judgement:} \textcolor{darkgreen}{Correct}\textcolor{red}{$\times$}
 \\
 \textbf{Reason:} The candidate's answer correctly identifies the location of Marino Railway Station and its opening date.
\end{tabular}  
&
\multirow{3}{\hsize}{ \begin{tabular}{>{}m{0.23\textwidth}}\textbf{Text: }Marino railway station is on the Seaford railway line , in the southern Adelaide suburb of Marino , South Australia Once the terminus of the line that now runs south to Seaford . \textbf{Marino railway station opened in 1915 when the line was built from Adelaide to Marino.} \end{tabular} } \\
\cline{1-2}

\textbf{\method} &
\begin{tabular}{>{\raggedright\arraybackslash}m{0.6\textwidth}}
 \textbf{Question:} What is the common trait of Marino railway station in Adelaide, …, and what years were they opened? \\ 
 \textbf{Answer:} Marino Railway Station is located in Adelaide, Australia. It was opened on October 19, 1885, and it is still operational. It serves the Seaford and Tonsley lines of Adelaide's metropolitan train network.\\
  \textbf{Judgement:} \textcolor{red}{Incorrect}\textcolor{darkgreen}{$\checkmark$}\\
 \textbf{Reason:} The candidate incorrectly states that Marino Railway Station was opened on October 19, 1885, while the reference text indicates it opened in 1915.
\end{tabular} & \\
\midrule
\midrule

\multicolumn{3}{l}{\cellcolor{mygray} \textbf{Performance of GPT3.5-Turbo}} \\
% \midrule
\textbf{Jw/R} &
\begin{tabular}{>{\raggedright\arraybackslash}m{0.6\textwidth}}
 \textbf{Question:} In legal terms, what does the term `damage' specifically refer to? \\
 \textbf{Answer:} In legal terms, the term ‘damage’ refers to the harm or injury caused to a person, property, or reputation as a result of the wrongful act of another… \\
 \textbf{Judgement:} \textcolor{red}{Incorrect}
\end{tabular} &
\multirow{3}{\hsize}{ \begin{tabular}{>{}m{0.23\textwidth}} \textbf{Text: }The word `damage' traditionally describes a harm to property (hence, `property damage'), rather than harm to the person (usually referred to as 'personal injury').\end{tabular}} \\
\cline{1-2}

\textbf{\method} &
\begin{tabular}{>{\raggedright\arraybackslash}m{0.6\textwidth}}
 \textbf{Question:} In area of law, …, and what is the distinction between `damage' and `injury’? \\
 \textbf{Answer:} The distinction between `damage' and `injury' is that damage refers to harm caused to property, while injury refers to the victim by certain injury. \\
 \textbf{Judgement:} \textcolor{darkgreen}{Correct}
\end{tabular} & \\

\cline{1-2}
\textbf{Human} &
\begin{tabular}{>{\raggedright\arraybackslash}m{0.6\textwidth}}
 \textbf{Question:} In legal terminology, what kind of object does `damage' apply to ? \\
 \textbf{Answer:} `Damage' usually applies to damage to objects. This includes loss or damage to property, land, equipment, etc… \\
 \textbf{Judgement:} \textcolor{darkgreen}{\textcolor{darkgreen}{Correct}}
\end{tabular} & \\

\bottomrule
\end{tabularx}
\caption{Comparison of \method~and Jw/oR at the top of the table, and performance of GPT3.5-Turbo of different baselines at the bottom of the table.
}
\label{tab:case}
\end{table*}
\subsection{Evaluation Results}

As shown in Tab.\ref{tab:main1}, \textbf{LLMs demonstrate superior performance across diverse knowledge domains, with GPT-4 consistently achieving the highest accuracy scores among all tested models.}
GPT-4 achieves remarkable accuracy rates of 95.4\% on Code, 94.6\% on Legal.term, and 97.9\% on Med.sim datasets.
These outstanding results underscore its versatility in handling both general and specialized tasks, particularly in biomedical domains. 
When compared to GPT-3.5, which achieves 90.8\% on Legal.term and 54.5\% on Wiki, the performance gap clearly illustrates the advantages of larger-scale models in knowledge-intensive tasks.

As shown in Tab.\ref{tab:main1}, \textbf{domain-specific fine-tuning yields substantial improvements in model performance within targeted fields.}
This is evidenced by PMC-LLaMA's performance in the biomedical domain, where it achieves accuracy scores of 78.5\% and 75.7\% on Med.rand and Med.sim, respectively.
While these scores don't surpass GPT-4's performance, they represent a significant improvement over general-purpose models like LLaMA2, which only achieves 73.2\% and 72.5\% on the same datasets.

\subsection{Evaluation Performance with Reference Data}

As shown in Tab.\ref{tab:wo}, \textbf{models exhibit a bias toward providing `correct' judgments when evaluating without external references.}
While Jw/oR assigns consistently higher scores to GPT-3.5 compared to \method, manual verification reveals that \method~actually achieves higher evaluation accuracy. 
This discrepancy suggests that Jw/oR's inflated results stem from its tendency to classify wrong answers as correct incorrectly.

As shown in Tab.\ref{tab:performance_ref_eval_uncommon}, \textbf{reference-based evaluation through \method~ensures objective assessment by strictly adhering to provided reference materials, even when they conflict with real-world knowledge.} 
This is exemplified in the case where GPT-4's response about a film ``winning the Golden Rooster Award for Best Directorial Debut in 2005'' was marked incorrect due to contradicting reference information, despite being factually accurate.
In contrast, using Jw/oR, GPT-4 achieved high accuracy on the Uncommon dataset by relying on its internal knowledge.
Manual verification confirms \method's strict adherence to reference materials in over 90\% of cases, regardless of potential conflicts with the model's internal knowledge base.

\begin{figure*}[t]
    \centering
    \begin{subfigure}{0.49\textwidth}
        \centering
        \includegraphics[width=\textwidth]{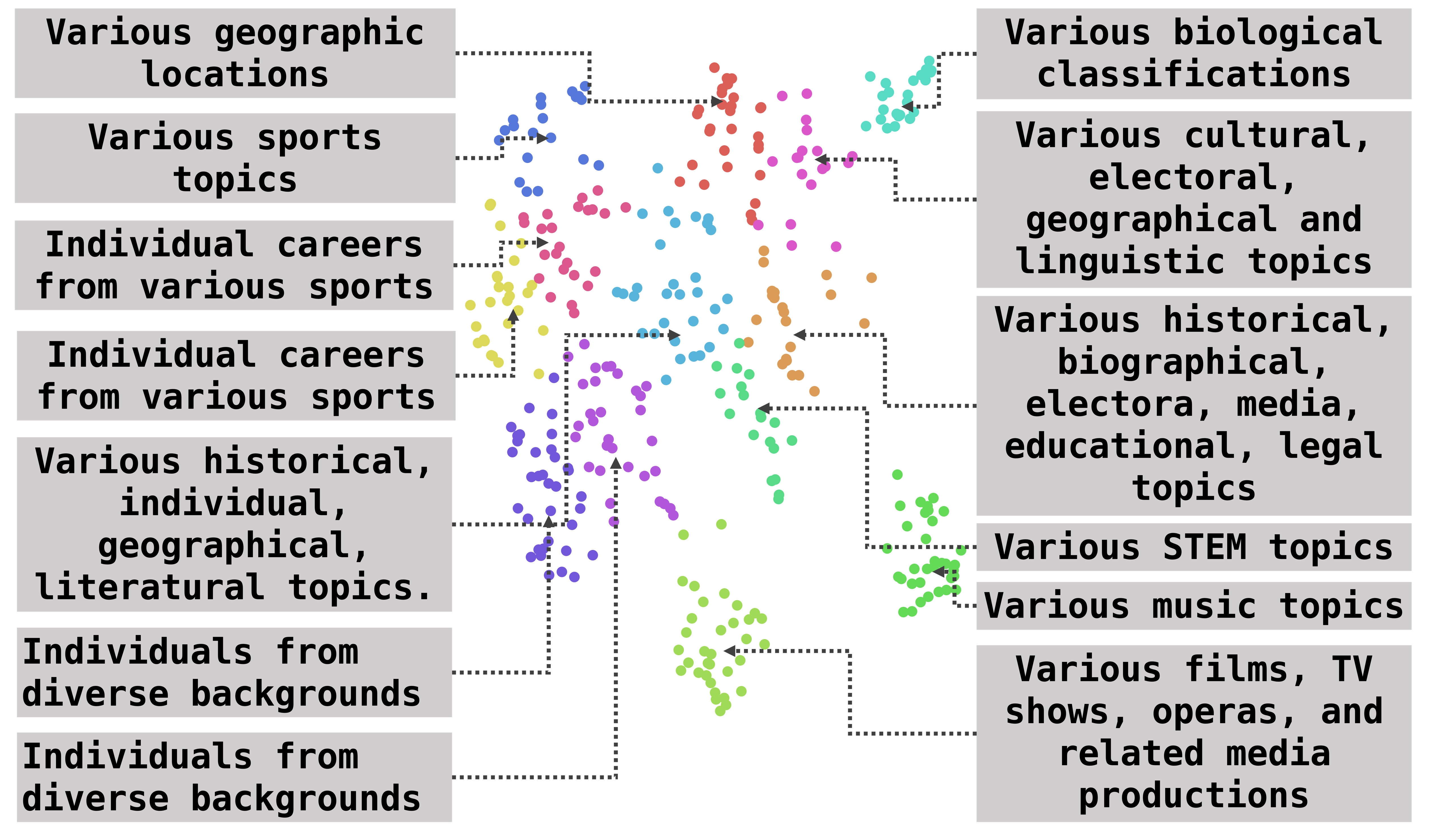}
        \caption{k=30}
        \label{fig:subfig1}
    \end{subfigure}
    \hfill
    \begin{subfigure}{0.49\textwidth}
        \centering
        \includegraphics[width=\textwidth]{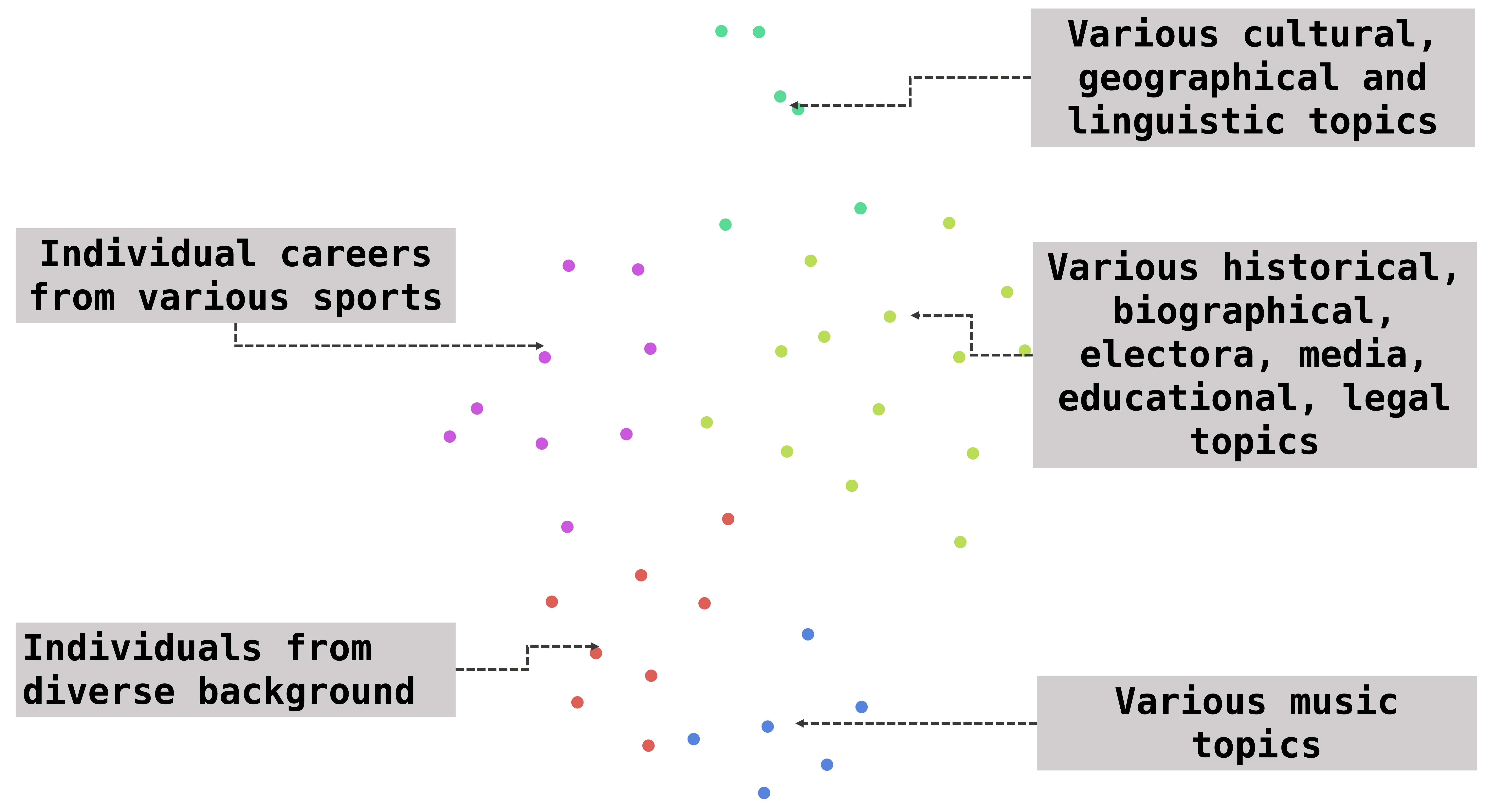}
        \caption{k=40}
        \label{fig:subfig2}
    \end{subfigure}
    \caption{The embedding map of the remaining knowledge units to be evaluated on the Wiki dataset in the process of evaluating GPT3.5-turbo at round \textbf{k} (k is different of the two sub-figures), where the representation of knowledge units of the same color is clustered in the same class, and their class names are displayed.}
    \label{fig:cluster}
\end{figure*}

\begin{figure}[t]
    \centering
    \includegraphics[width=0.99\columnwidth]{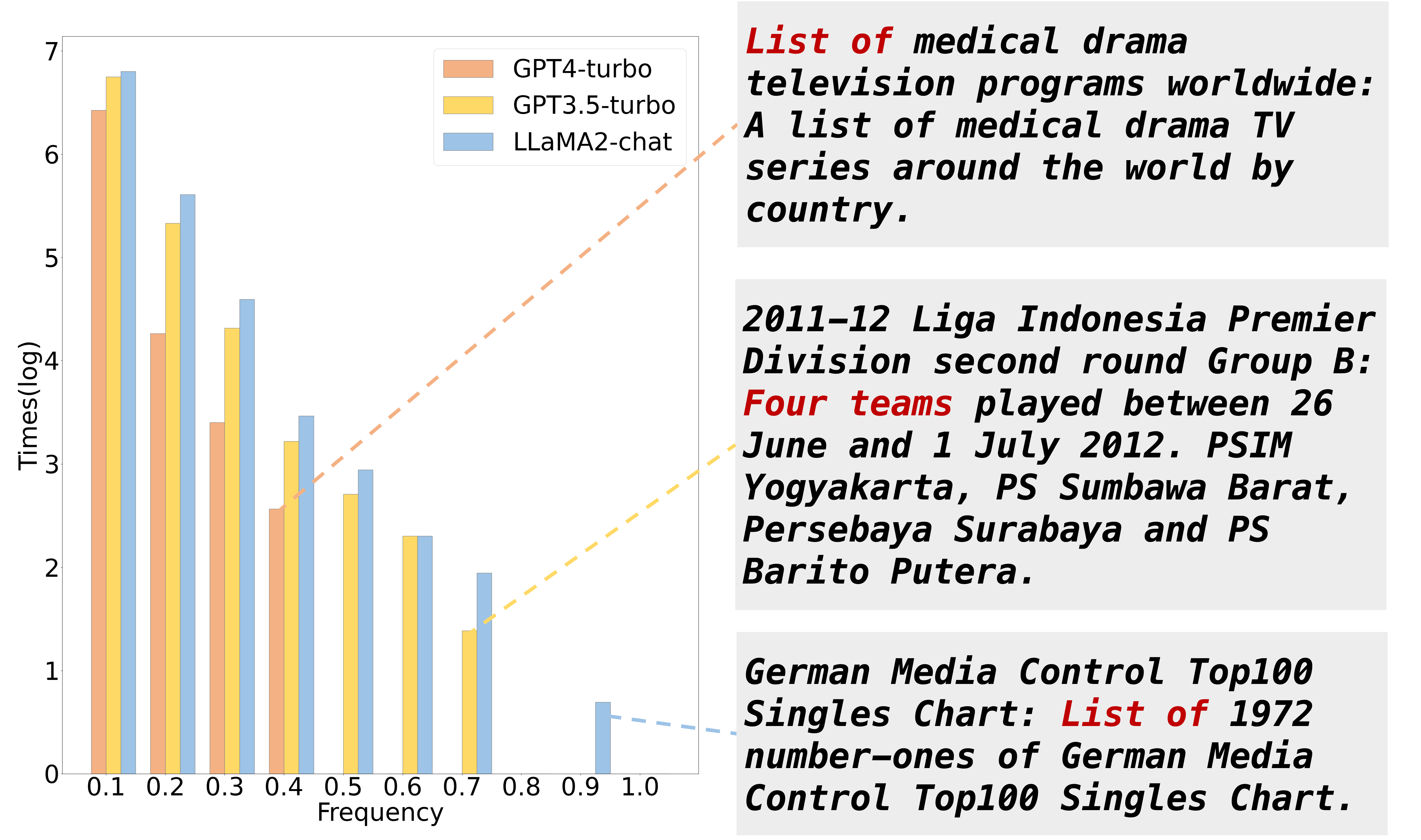}
    \caption{The frequency of knowledge units in multiple iterations of \method. The X-axis is the frequency of knowledge units, and the vertical axis is the number of knowledge units of a certain frequency.}
    \label{fig:frq_time}
\end{figure}

\subsection{Evaluation Performance Knowledge Clustering and Concept Characteristics}

As shown in Figure \ref{fig:subfig1}, \textbf{knowledge clusters with high intra-cluster similarity and low inter-cluster overlap optimize \method's evaluation accuracy.}
The analysis reveals 13 distinct clusters characterized by high intra-cluster semantic similarity and large inter-cluster distances in the feature space.
Between rounds 30 (Figure \ref{fig:subfig1}) and 40 (Figure \ref{fig:subfig2}), domain-specific clusters such as ``Various films'', ``TV shows'', ``operas'', and ``related media productions and Various STEM topics'' achieve complete evaluation coverage. 
In contrast, heterogeneous clusters containing mixed domains, like \textit{``Various historical, biographical, electoral, media, educational and legal topics''}, exhibit high feature dispersion in the embedding space, leading to reduced evaluation precision.

As shown in Fig.\ref{fig:frq_time}, \textbf{the hierarchical clustering strategy in \method~optimizes evaluation efficiency through systematic knowledge unit distribution.}
The frequency analysis demonstrates an inverse relationship between knowledge unit sampling frequency and the occurrence of `ignored' units.
This pattern validates that \method's clustering algorithm effectively prevents redundant sampling of high-frequency knowledge units while maintaining comprehensive coverage across the knowledge space.

As shown in Fig.\ref{fig:frq_time}, \textbf{LLMs' performance exhibits systematic degradation when processing knowledge units with high information density.} 
This is evidenced by the increased frequency of `ignored' knowledge units containing extensive information, such as \textit{``List of medical drama television programs worldwide''}. 
The performance degradation is particularly notable in sparse clusters, suggesting that LLMs' processing efficiency decreases as the information density of the input content increases.

\subsection{Evaluation Performance of Different Types of Questions}

As shown in Fig.\ref{fig:question_classification}, \textbf{LLMs consistently exhibit lower performance on identity-based and location-based queries.} 
The data reveals a systematic underperformance across all models when handling `who' questions, likely due to the complexities of processing person-specific information and historical contexts. 
Similar performance deficits are observed with `where' questions, suggesting challenges in managing geographical details and location-based information. 
This pattern indicates fundamental limitations in models' ability to process and accurately represent identity and location-specific knowledge.

As shown in Tab.\ref{tab:case}, \textbf{\method~enhances model performance by leveraging contextual information to activate inherent reasoning capabilities.} While GPT-4-turbo and LLaMa2-chat show superior performance with Jw/R on factual recall questions ('who' and `where'), \method~achieves comparable or better results with models like GPT-3.5-turbo by incorporating rich contextual information. For instance, \method~improves GPT-3.5-turbo's understanding of `damage' concepts by providing relevant contextual cues, whereas Jw/R struggles without such supporting information.

\subsection{Case Study}

As shown in Tab.\ref{tab:corr}, \textbf{\method~demonstrates superior evaluation accuracy compared to Jw/oR through its reference-based assessment approach.} 
This is exemplified by the case of the Marino Railway Station opening time query, where Jw/oR incorrectly marks an answer as `Correct' based on the model's internal knowledge, while \method~accurately identifies it as `Incorrect' by comparing it against the Wiki reference. 
This case highlights \method's ability to provide reliable evaluations through reference-based verification rather than depending solely on model knowledge.

\section{Conclusion}
In this paper, we introduce \method, a novel method for evaluating large language models across diverse domains. 
Leveraging LLMs’ ability to understand and respond to complex queries, \method~addresses the limitations of traditional evaluation methods. 
Our experiments on various datasets show that \method~achieves high consistency with human evaluation, which highlights its effectiveness in assessing model responses. 
This approach not only offers a scalable and efficient means to evaluate LLMs but also advances the field of model evaluation in knowledge-intensive tasks.

\section*{Limitation}
While the proposed \method~offers significant advancements in evaluating text generated by language models, it is not without limitations. 
\method's effectiveness heavily depends on the quality and relevance of the external reference materials used. If these references are incomplete or outdated, the framework's evaluations may be compromised. The challenge of integrating large volumes of reference data remains, as even with synthesized knowledge units, the risk of overlooking critical details or context persists.  
The sensitivity of LLMs to prompt variations can also result in inconsistent evaluation outcomes when dealing with diverse or ambiguously phrased questions. This variability in model responses may affect the reliability of the evaluation results, particularly in scenarios where nuanced understanding is crucial. 

\section*{Ethical Concerns}
The \method~introduces several ethical concerns. The external reference data used may include sensitive or controversial content, which could lead to the perpetuation of biases or misinformation. Additionally, handling proprietary or personal information raises privacy and intellectual property concerns.

To address these issues, we implement strict protocols to vet reference data for sensitivity and relevance. We ensure transparency in our data curation process and prioritize ethical standards to safeguard privacy and prevent misuse, balancing the benefits of comprehensive evaluation with responsible data handling.

\bibliography{custom}

\appendix

\section{Appendix}
\label{sec:appendix}
\begin{table*}
\centering
\begin{tabular}{c|ccccc}
\hline
Model & Score 1 & Score 2 & Score 3 & Score 4 & Score 5 \\
\hline
GPT-4 & 0.9566 & 0.9479 & 0.8269 & 0.9190 & 0.9230 \\
GPT-3.5 & 0.9245 & 0.9056 & 0.5094 & 0.7735 & 0.8113 \\
Llama2 & 0.8219 & 0.8062 & 0.4002 & 0.6295 & 0.6352 \\
MedLlama & 0.6367 & 0.6241 & - & - & - \\
\hline
\end{tabular}
\caption{Human scores for different models}
\label{tab:human_scores}
\end{table*}

\subsection{Detials of \method}
\label{A.1}
\subsubsection{Prompt}
\label{A.1.1}
\begin{figure*}
    \centering
    \includegraphics[width=0.8\linewidth]{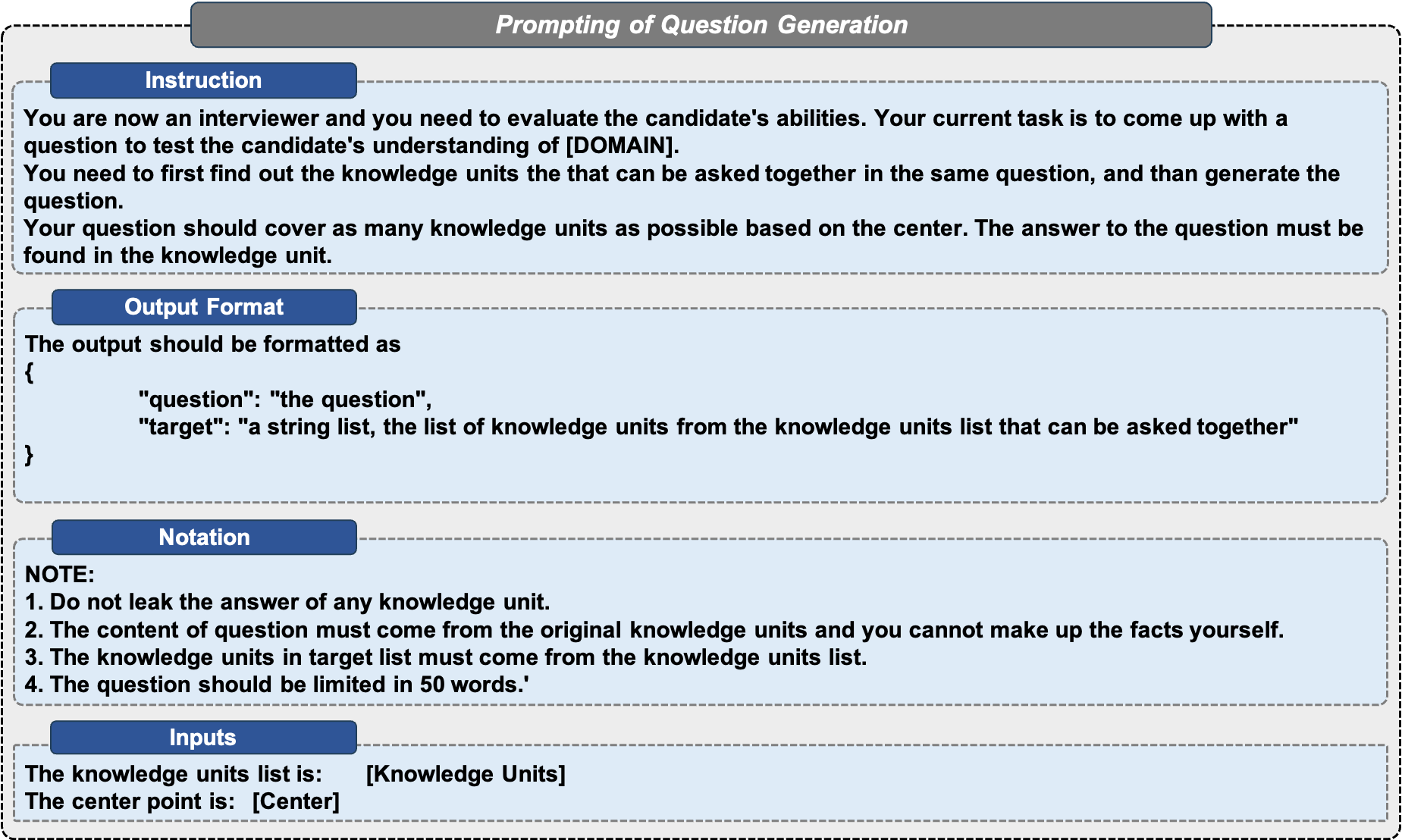}
    \caption{Prompt for question generation.}
    \label{fig:prompt_qg}
\end{figure*}

\begin{figure*}
    \centering
    \includegraphics[width=0.8\linewidth]{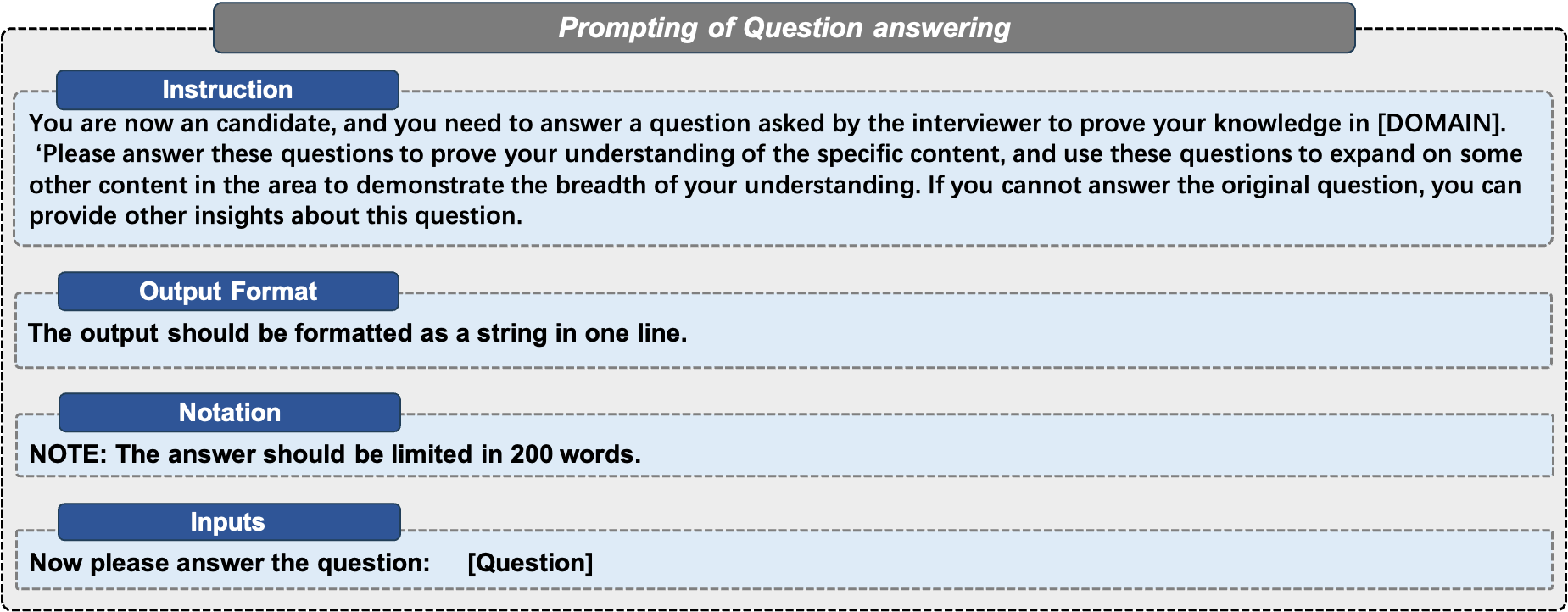}
    \caption{Prompt for question answering.}
    \label{fig:prompt_qa}
\end{figure*}

\begin{figure*}
    \centering
    \includegraphics[width=0.8\linewidth]{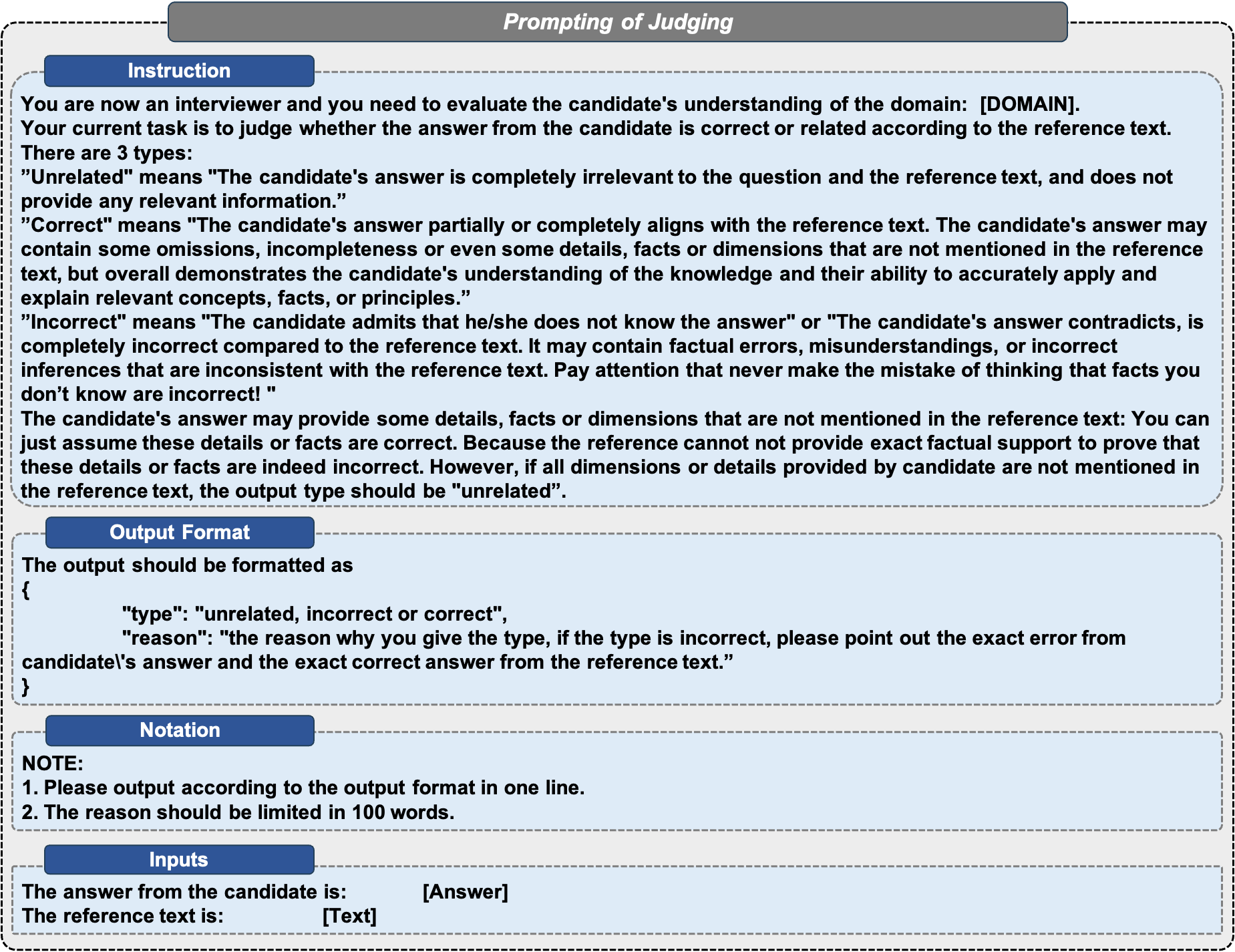}
    \caption{Prompt for response judgment.}
    \label{fig:prompt_jd}
\end{figure*}

In Fig.\ref{fig:prompt_qg}, Fig.\ref{fig:prompt_qa}, and Fig.\ref{fig:prompt_jd}, we illustrate three distinct prompts, each designed to complete different tasks within \method.
Fig.\ref{fig:prompt_qg} represents a question generation prompt in step 3 of \textbf{Question Generation}. 
Fig.\ref{fig:prompt_qa} represents a question-answering prompt in step 4 of \textbf{Get Response}.
Fig.\ref{fig:prompt_jd} represents a response judgment prompt in step 5 of \textbf{Judging}.

Each of them includes an instruction, an output format, a notation, and inputs.

\subsubsection{Details}
\label{A.1.2}
\begin{table*}[ht]
\begin{center}
\caption{The models, inputs, and outputs of each step in \method}
\label{tab:detail}
% 第一部分表格
\resizebox{0.99\textwidth}{!}{
\begin{tabular}{@{}c|c|c|c@{}}

\toprule
Step & Model & Input Example & Output Example \\ \midrule

\begin{tabular}{>{\raggedright\arraybackslash}m{0.1\textwidth}}Preparing Knowledge Unit Candidates\end{tabular}
& \begin{tabular}{>{\raggedright\arraybackslash}m{0.07\textwidth}}gpt3.5-turbo-0125 \end{tabular}
&\begin{tabular}{>{\raggedright\arraybackslash}m{0.39\textwidth}}
\textbf{"text": }"lastaired = ' ' Sa Sandaling Kailangan Mo Ako ' ' ( lit . ' ' The Moment You Need Me ' ' ) is a Philippine drama series aired on ABS-CBN in 1998-2001 . It starred some of the Philippines ' renowned veteran actors and actresses combined with promising young artists under the direction of the reputable Filipino TV and movie director , Olivia M. Lamasan . This shows the intricacies of love , friendship , betrayal and forgiveness. \# Cast \# *Marvin Agustin as Ruben *Kristine Hermosa as Agnes *Piolo Pascual as Raffy *Giselle Toengi as Stella *John Lloyd Cruz as Daniel *Kaye Abad as Eloisa *Dante Rivero *Hilda Koronel *Ronaldo Valdez *Tessie Tomas *Tracy Vergel *William Lorenzo \# See also \# *List of programs broadcast by ABS-CBN *List of programs aired by ABS-CBN *List of telenovelas of ABS-CBN"
\end{tabular}
& \begin{tabular}{>{\raggedright\arraybackslash}m{0.39\textwidth}}
\textbf{"type"}: "Factual Knowledge", \\
\textbf{"keyword"}: "Sa Sandaling Kailangan Mo Ako", \\
\textbf{"description"}: "A Philippine drama series aired on ABS-CBN from 1998-2001, directed by Olivia M. Lamasan."
\end{tabular}
\\ \midrule

\begin{tabular}{>{\raggedright\arraybackslash}m{0.1\textwidth}}Clustering\end{tabular}       
& \begin{tabular}{>{\raggedright\arraybackslash}m{0.07\textwidth}}gpt4-turbo-0409 \end{tabular}
&\begin{tabular}{>{\raggedright\arraybackslash}m{0.39\textwidth}}
\textbf{"knowledge\_units"}: ["bundle to pass data: Passing required data to subfolders using bundle.", "Types of storage in Android: Shared Preferences, Internal storage, External storage, SQLite database, Network connection", "Service: A background component for running music or processing network transactions", "Fragment: A part of an Activity to display multiple screens.", "Sleep mode for CPU: CPU is in a sleep state and not accepting commands except for wireless and alarms.", ...(and 729 more)]
\end{tabular}          
&\begin{tabular}{>{\raggedright\arraybackslash}m{0.39\textwidth}}
\textbf{"center"}: "Web Services: Open standards for cross-platform communication between applications",
\textbf{"knowledge\_units"}: ["Web services: Client and server apps communicate with HTTP protocol for interoperability.", "SOAP and REST: SOAP for data transmission, REST for handling media components.", "REST and Routing in Rails: REST represents representational state transfer, uses HTTP methods to represent actions", "API and Web Services differences: APIs may not require network access, while Web Services always do", "SOAP: XML-based protocol to exchange information between computers.", "REST API: A set of functionalities to perform requests and responses over HTTP.", "REST features: REST is stateless, uses GET method to access resources, and differs from web services.", ...(and 64 more)]\\
...(more centers and knowledge\_units)
\end{tabular}   

\\ \midrule

\end{tabular}
}

\end{center}
\end{table*}
\newpage
\begin{table*}[ht]
\begin{center}
% 第2部分表格
\resizebox{0.99\textwidth}{!}{
\begin{tabular}{@{}c|c|c|c@{}}

\midrule
\begin{tabular}{>{\raggedright\arraybackslash}m{0.1\textwidth}}Question Generation\end{tabular}
&\begin{tabular}{>{\raggedright\arraybackslash}m{0.07\textwidth}}gpt4-turbo-0409 \end{tabular}
&\begin{tabular}{>{\raggedright\arraybackslash}m{0.39\textwidth}}
\textbf{"center"}: "Exception handling in various programming languages", 
\textbf{"knowledge\_units"}: ["Exception handling: Mechanism used to handle runtime errors and maintain normal flow of program.", "Exception Handling: Process of catching and handling exceptions in a program", "Exception Handling in Java: How catch block changes variable value and how exception handling works", "Exception handling in Django: Using exception classes in Django to handle anomalous events", "Exception Handling: Begin-end block with rescue clause specify exception type to handle.", "Exception handling: A mechanism that handles exceptional situations and prevents abnormal program termination.", "Quit command: A command to close the connection.", "rmdir command: Removes the directory specified by the filename.", "Python exception handling: Using try-except blocks to handle errors, with available clauses try-except-finally and try-except-else.", "Exception handling: A mechanism to deal with runtime errors, maintaining the normal program flow.", "Exception handling in Java: Handling and propagation of exceptions in Java programs", "System class: Access system resources, standard input and output.", "gunzip command: a command used to decompress gzip files", "Exception handling in Django: Django uses its own exception classes and supports all core Python exceptions.", "Exception Handling: Using begin-end and rescue clause to handle exceptions", "Debugging: The process of finding and fixing errors in code.", "'type' and 'typeAndWait' command: Enter keyboard key or select value, wait for page reload", ...(and 10 more)]
\end{tabular}
&\begin{tabular}{>{\raggedright\arraybackslash}m{0.39\textwidth}}
\textbf{"question"}: "In the context of various programming languages, explain how exception handling works in Java, including how a catch block can change a variable's value, the use of try-except blocks in Python with their clauses, and how Django utilizes its own exception classes to handle errors. Additionally, contrast these mechanisms with the general process of catching and managing exceptions to maintain the normal flow of a program. Then tell me what else you know about these.", "center": "Exception handling in various programming languages"
\end{tabular}
\\ \midrule

\end{tabular}
}
\end{center}
\end{table*}
\newpage
\begin{table*}[ht]
\begin{center}
% 第2部分表格
\resizebox{0.99\textwidth}{!}{
\begin{tabular}{@{}c|c|c|c@{}}

\midrule
\begin{tabular}{>{\raggedright\arraybackslash}m{0.1\textwidth}}Get Response \end{tabular}     
& \begin{tabular}{>{\raggedright\arraybackslash}m{0.07\textwidth}}model to be evaluated\end{tabular}
&\begin{tabular}{>{\raggedright\arraybackslash}m{0.39\textwidth}}
\textbf{"question"}: "In Java object-oriented programming, how does the concept of inheritance differ from implementing an interface, and can you explain the use of the 'javap' command in relation to these concepts? Also, can you describe the role of JavaBeans in encapsulating state and behavior within an object, the differences between singleton and prototype bean scopes in Spring, and how thread creation can be achieved? Please include an explanation of how a thread can be interrupted and the implications of calling 'join()' on a thread, along with the considerations for thread-safety when dealing with singleton patterns. Then tell me what else you know about these."
\end{tabular}
&\begin{tabular}{>{\raggedright\arraybackslash}m{0.39\textwidth}}
\textbf{"answer"}: "In Java, inheritance is a mechanism where one class acquires the properties (methods and fields) of another class. The child class extends the parent class, and it can use all the public and protected methods and fields of the parent. On the other hand, implementing an interface provides a way for a class to become more formal about the behavior it promises to provide. It specifies a set of methods that the class must implement. Unlike inheritance, a class can implement multiple interfaces, but it can only extend one class.  The 'javap' command in Java is used to print information about the classes and interfaces in a given Java binary file. It displays the package, protected, and public fields and methods of the classes available in the specified file. JavaBeans are reusable software components that adhere to specific naming conventions for properties, methods, and events. They provide a way to encapsulate state and behavior within an object. Singleton and prototype bean scopes are used in Spring to control the lifecycle of objects. Singleton beans are created only once per container, while prototype beans are created every time they are requested.  Thread creation in Java can be achieved by either extending the Thread class or implementing the Runnable interface. Interrupting a thread is done by calling the interrupt() method of the thread object. Calling join() on a thread waits for it to complete before moving on to the next task. Thread-safety considerations are important when dealing with singleton patterns because multiple threads may access the same instance concurrently.  In addition to these topics, I have knowledge of other Java concepts such as generics, collections, exception handling, JDBC, JPA, Hibernate, Servlets, JSP, RESTful web services, Spring Framework, and more."
\end{tabular}
\\ \midrule
\end{tabular}
}\end{center}
\end{table*}
\newpage
\begin{table*}[ht]
\begin{center}
% 第2部分表格
\resizebox{0.99\textwidth}{!}{
\begin{tabular}{@{}c|c|c|c@{}}

\midrule
\multirow{2}{*}
{\begin{tabular}{>{\raggedright\arraybackslash}m{0.1\textwidth}}Judging\end{tabular}}              
& \begin{tabular}{>{\raggedright\arraybackslash}m{0.07\textwidth}}gpt3.5-turbo-0125\end{tabular}
& \begin{tabular}{>{\raggedright\arraybackslash}m{0.39\textwidth}}
\textbf{"knowledge\_unit"}: "Java Bean: A reusable software component that encapsulates many objects into one object for easy maintenance.",\\
\textbf{"answer"}: "In Java, inheritance is a mechanism where one class acquires the properties (methods and fields) of another class. The child class extends the parent class, and it can use all the public and protected methods and fields of the parent. On the other hand, implementing an interface provides a way for a class to become more formal about the behavior it promises to provide. It specifies a set of methods that the class must implement. Unlike inheritance, a class can implement multiple interfaces, but it can only extend one class.   The 'javap' command in Java is used to print information about the classes and interfaces in a given Java binary file. It displays the package, protected, and public fields and methods of the classes available in the specified file. ...(more)"
\end{tabular}            
&  \begin{tabular}{>{\raggedright\arraybackslash}m{0.39\textwidth}}
\textbf{"knowledge\_unit related content"}: "JavaBeans are reusable software components that adhere to specific naming conventions for properties, methods, and events. They provide a way to encapsulate state and behavior within an object. Singleton and prototype bean scopes are used in Spring to control the lifecycle of objects. Singleton beans are created only once per container, while prototype beans are created every time they are requested."
\end{tabular}                    
\\ \cmidrule(l){2-4}                                                                   
                                                                              
&\begin{tabular}{>{\raggedright\arraybackslash}m{0.07\textwidth}}gpt4-turbo-0409 \end{tabular}
& \begin{tabular}{>{\raggedright\arraybackslash}m{0.39\textwidth}}
\textbf{"knowledge\_unit related content"}: "JavaBeans are reusable software components that adhere to specific naming conventions for properties, methods, and events. They provide a way to encapsulate state and behavior within an object. Singleton and prototype bean scopes are used in Spring to control the lifecycle of objects. Singleton beans are created only once per container, while prototype beans are created every time they are requested.", \\"text": "Singleton scope should be used together with EJB stateless session bean, and prototype scope should be used together with EJB stateful session bean."
\end{tabular}            
&  \begin{tabular}{>{\raggedright\arraybackslash}m{0.39\textwidth}}
\textbf{"type"}: "incorrect", \\
\textbf{"reason"}: "The candidate incorrectly associates singleton and prototype bean scopes with Spring, while the reference text specifically mentions their use with EJB (Enterprise JavaBeans) stateless and stateful session beans, not Spring."
\end{tabular}                       
\\ 
\bottomrule
\end{tabular}
}\end{center}
\end{table*}

The models, inputs, and outputs of each step are shown in Tab.\ref{tab:detail}.

\subsection{Dataset}
\label{A.2}
\textbf{Med.rand} consists of randomly chosen questions from PubMedQA. 
\textbf{Med.sim}, on the other hand, is a selection from PubMedQA based on the similarity to the paragraph: \textit{“A medical history of arterial hypertension was associated with lower MMSE scores and a higher prevalence of dementia and cognitive decline at baseline. However, intact cognition through the observation period was linked to higher baseline SBP.”} The similarity is determined by comparing the embeddings of this paragraph with those of all other paragraphs in PubMedQA.

\subsection{Human Baseline}
\label{A.3}
\subsubsection{Human Baseline Description}
\paragraph{Annotator Selection
}
We selected three annotators with expertise in the relevant fields to ensure the quality of the questions and annotations. All annotators had prior experience in data annotation and a good understanding of the subject matter.
\paragraph{Question Formulation}
The annotators were instructed to manually formulate questions based on the original text provided in the datasets. They were asked to create questions that would test the comprehension and response-generation capabilities of the models.
\paragraph{Annotation Process}
The annotators annotated the model responses while having access to the original text. This approach allowed them to assess the accuracy of the model's answers in the context of the given information.
\subsubsection{Annotation Scoring}
\paragraph{Scoring Criteria}
Annotations were scored on a binary scale: 0 for incorrect answers and 1 for correct answers. An answer was considered correct if it accurately addressed the question based on the information provided in the original text.

\paragraph{Scores}
The average scores for each dataset and model are presented in Tab.\ref{tab:human_scores}:

\paragraph{Inter-Annotator Agreement}
To ensure the reliability of the annotations, we calculated the inter-annotator agreement using the Fleiss' kappa coefficient. The kappa value was found to be 0.72, indicating substantial agreement among the annotators.

\subsection{Other Factors in \method}\label{sec:ana1}

\begin{figure*}[t]
    \centering
    \includegraphics[width=0.99\textwidth]{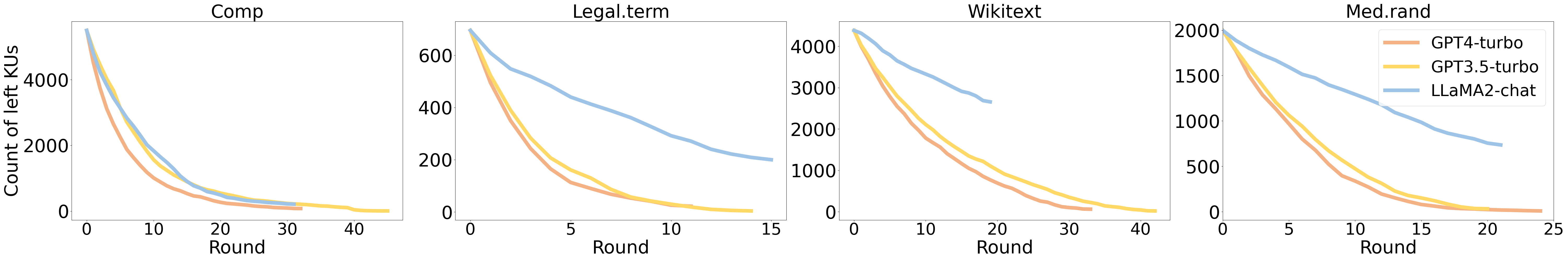}
    \caption{The number of unjudged knowledge points remaining after each iteration of different models on different data sets. The iteration ends when the representation on \method~converges.}
    \label{fig:model_cap}
\end{figure*}

\textbf{
The more knowledge units reduce in each chatting round, the better the LLM's ability to answer relevant knowledge questions.}
% \gzh{Is this statement the same as the statement in the next paragraph? I feel token expense is lower because knowledge units reduce faster.}
Fig.\ref{fig:model_cap} shows the performance of LLM across multiple iterations of \method.
The slope of these curves, denoted as the difference between the count of knowledge units in the \((n+1)\)th round and the \(n\)th round, represents the amount of knowledge units that are reduced in each round.
More powerful models like GPT4-turbo tend to answer more questions in each chatting round, especially at the outset and particularly on challenging datasets.
While LLaMA2-Chat consistently exhibits slower speeds compared to GPT4-turbo and GPT3.5-turbo, which indicates the low capacity of LLaMA2-Chat for question answering compared to the other models.
% \gzh{This paragraph may put into the appendix}

\textbf{In \method, Models with stronger capacity incur fewer token costs during the evaluation process in our method.}
% \gzh{Still feel strange about your statement, still think you didn't state it clearly.}
Costs of GPT4-turbo and LLaMA2-chat in Tab.\ref{tab:effi} reflect that the cost of evaluating GPT4 on \method~is 26\% lower than that of LLaMA. 
This cost advantage shows in the stage of both Question Generstion and Response Judgement. 
As shown in Fig.\ref{fig:model_cap}, the amount of knowledge units that are reduced in each round is different between models, and this also causes the difference in cost between models. 
Specifically, GPT4-turbo has a significant reduction in the number of knowledge units in each round, which decreases the repetition of questions and judgments, thereby reducing the cost of API calls.
% \gzh{This paragraph may put into the appendix}

% \end{CJK}
\end{document}